\documentclass{article}

\usepackage[preprint]{neurips_2023}

\usepackage[utf8]{inputenc}
\usepackage[T1]{fontenc}
\usepackage{hyperref}
\usepackage{url}
\usepackage{booktabs}
\usepackage{amsfonts}
\usepackage{amsmath}
\usepackage{amssymb}
\usepackage{nicefrac}
\usepackage{microtype}
\usepackage{xcolor}
\usepackage{graphicx}
\usepackage{doi}
\usepackage{tikz}
\usepackage{pgfplots}
\pgfplotsset{compat=1.18}
\usetikzlibrary{shapes,arrows,positioning,calc,backgrounds,fit,patterns,decorations.pathreplacing}
\usepackage{amsthm}

\hypersetup{
  colorlinks=true,
  linkcolor=[rgb]{0.1,0.2,0.55},
  citecolor=[rgb]{0.1,0.2,0.55},
  urlcolor=[rgb]{0.1,0.2,0.55}
}

\makeatletter
\renewcommand{\@notice}{}
\makeatother

\title{Measuring Cross-Task Behavioral Consistency in Language Model Agents}

\author{%
  Amritesh Banerjee\footnotemark[1] \\
  University of Massachusetts Amherst \\
  \texttt{amriteshbane@umass.edu} \\
  \And
  Pranil Raichura\thanks{Equal contribution.} \\
  Mantis AI Research (MIT CSAIL) \\
  \texttt{raichura@mit.edu} \\
}

\begin{document}
\maketitle

\begin{abstract}
Agent evaluation relies almost entirely on outcome metrics such as success
rate, which capture whether an agent succeeds but not how consistently it
behaves. We argue that behavioral consistency across tasks is a distinct and
measurable property, and we introduce the Behavioral Consistency Metric (BCM)
to quantify it. BCM trains a model to predict task success from behavioral
features of agent execution traces, derives a per-trajectory feature-attribution
vector, and measures the mean pairwise similarity of these vectors within an
agent system. Across roughly 9{,}000 trajectories from six language model agents
on software engineering tasks, our central finding is that cross-task and
within-task consistency are distinct axes that can diverge: some systems are
locally reproducible, behaving similarly on repeated attempts at one task, yet
globally fragmented, with no stable strategy across different tasks, while others
are consistent at both scales. Prior work measures only same-task
reproducibility and so cannot observe this separation. We further find that
consistency is not reducible to success rate, since systems with comparable
success can differ sharply in consistency, and that the frontier-versus-open-source
consistency gap persists under a within-task control that holds task difficulty
constant. We position BCM as a process-level reliability signal that complements
outcome metrics, and we are explicit about the conditions under which it is
meaningful.
\end{abstract}

\vspace{0.3em}

\section{Introduction}

Language model agents are increasingly evaluated the way static models are: by
how often they produce the correct final result. For agents that solve tasks
through long sequences of actions, the dominant metrics are outcome metrics such
as success rate and pass@k, which record whether a task was resolved but say
nothing about how the agent arrived there. This is an efficient way to rank
systems, and it has driven rapid progress, and recent benchmarks formalize it
across many environments \citep{liu2024agentbench}. It also has a structural
blind spot, and a growing body of work argues that this outcome focus
systematically under-measures the properties that determine deployment value
\citep{jafarimeimandi2025measurement}. Two agents can resolve the same fraction
of tasks while behaving in fundamentally different ways, and an outcome metric
treats them as equivalent.

The distinction matters most for reliability. Two agents can each resolve about a
quarter of their tasks while one applies a stable, recognizable approach to every
task and the other reaches the same rate through a scattered collection of
approaches. By success rate alone they are indistinguishable, yet for anyone
deciding whether to deploy an agent the difference is decisive: a system that
behaves consistently can be characterized, anticipated, and trusted within known
bounds; one that behaves erratically cannot, even at identical average
performance.

We argue that this property, the consistency of an agent's behavior across
different tasks, is distinct from its success rate, is currently underexamined,
and can be measured directly from execution traces. Most existing work on agent
consistency studies a different question: whether an agent reproduces its
behavior when run on the same task more than once. That same-task reproducibility
is important, but it is not the same as whether an agent carries a coherent
strategy from one task to a different task. We call the latter cross-task
behavioral consistency, and it is the focus of this paper.

To measure it, we introduce the Behavioral Consistency Metric (BCM). We extract
structural features from each agent trajectory, such as how many steps it takes,
how often it errors, and how it allocates actions across searching, viewing, and
editing. We train a model to predict task success from these features, and for
each trajectory we derive an attribution vector describing which behavioral
features drove that prediction. These attribution vectors serve as a compact
behavioral representation of each run, and BCM is defined as their mean pairwise
similarity within an agent system. An agent that approaches different tasks
through a consistent strategy yields similar representations and a high BCM; an
agent whose behavior fragments across tasks yields dissimilar representations and
a low BCM. The attribution-vector representation follows prior work that uses such
vectors to cluster agent policies in other domains; our contribution is its
adaptation to measure cross-task consistency in language model agents.

Applied to roughly 9{,}000 trajectories from six agent systems on software
engineering tasks, BCM reveals structure that success rate conceals: cross-task
and within-task consistency emerge as distinct axes that can diverge, the
frontier-versus-open-source gap survives a within-task difficulty control, and
consistency comes apart from success rate.

To summarize our contributions: (i) we introduce BCM, a measure of cross-task
behavioral consistency for language model agents; (ii) we show that cross-task
and within-task consistency are distinct axes that can diverge, with some systems
locally reproducible yet globally fragmented, a separation invisible to the
same-task consistency measures used in prior work; (iii) we show that the
frontier-versus-open-source consistency gap persists under a within-task control
that holds task difficulty constant; and (iv) we show that consistency is not
reducible to success rate, since systems at comparable success can differ sharply
in consistency. We position BCM as a process-level reliability signal that
complements, rather than replaces, outcome metrics, and we are explicit about the
conditions under which it is meaningful.

\section{Related Work}

\paragraph{Beyond outcome metrics for agents.}
A growing body of work argues that success rate alone is an insufficient lens on
agent behavior, particularly for coding agents on SWE-bench. Several studies
analyze execution trajectories to distinguish successful from failed runs,
identifying patterns such as redundant exploration, premature fixes, and the
balance between information gathering and editing
\citep{majgaonkar2025understanding, bouzenia2025understanding}, and others build
diagnostic decompositions to predict model-level performance
\citep{kim2026trajeval}. Most relevant to us, recent work shows that the reported
relationship between trajectory length and failure largely reverses once task
difficulty is controlled, establishing that length reflects difficulty rather than
strategy quality \citep{mehtiyev2026beyond}; we take this confound seriously and
build a within-task control around it. These studies are descriptive or driver
analyses: they characterize what behaviors accompany success and failure, but none
defines a consistency measure or asks whether behavior is stable across tasks.
More broadly, agent benchmarks remain organized around task completion
\citep{liu2024agentbench}, and recent surveys document how this outcome focus
leaves process-level reliability under-measured \citep{jafarimeimandi2025measurement}.

\paragraph{Consistency and reliability of agents.}
A second line of work measures agent consistency directly, but almost exclusively
in the same-task sense: whether an agent reproduces its behavior or outcome when
run repeatedly on an identical task. This includes work measuring divergence
across repeated runs of the same agent on the same task \citep{mehta2026disagree}
and outcome-consistency metrics such as pass\textsuperscript{k}, the probability
that an agent succeeds on all of $k$ identical trials \citep{yao2025taubench}.
Closest in spirit, recent work studies behavioral consistency on the same
benchmark we use and reports that same-task consistency tends to track accuracy
across models \citep{mehta2026amplifies}. Our work differs along the axis that
defines it: rather than asking whether an agent repeats itself on one task, we ask
whether it carries a coherent strategy across different tasks, and we show these
are distinct properties, an agent can be consistent within a task yet fragmented
across tasks. This reconciles an apparent tension with prior findings: where
same-task consistency tracks accuracy, our cross-task measure dissociates from it,
which is consistent rather than contradictory once the two notions are separated.

\paragraph{Attribution vectors as behavioral representations.}
Our method represents each trajectory by the attribution vector of a success
predictor and compares these vectors by similarity. Using feature-attribution
vectors as a representation for clustering or comparing agent behavior is an
established technique in other domains, notably for clustering policies from
anonymous state-action data \citep{coletta2023kshap} and for analyzing learned
policies in reinforcement learning. We adopt this representational idea rather
than introduce it; our contribution is its application to measuring cross-task
behavioral consistency in language model agents, and the empirical findings that
application yields.

\section{Method}

\subsection{Problem Formulation}
The execution history of an agent operating in an environment is an ordered
sequence of action steps, $T = (a_1, a_2, \ldots, a_n)$, where each $a_i$ is a
single operational event such as a command execution or a file modification. To
compare trajectories across architectures, each is mapped to a fixed-length
feature space. For every trajectory we form a 12-dimensional structural feature
vector
\begin{equation}
\mathbf{x}_i = \left[ x_{i,1}, x_{i,2}, \ldots, x_{i,12} \right] \in \mathbb{R}^{12},
\end{equation}
whose components summarize execution depth, action length, repository
navigation, code-editing intensity, testing behavior, action diversity, local
repetition, error frequency, and action-category transitions; the full
definitions appear in Table~\ref{tab:features}.

From a success predictor (Section~\ref{sec:model}) we derive, for each
trajectory, an out-of-fold attribution vector
$\boldsymbol{\phi}_i \in \mathbb{R}^{12}$ that decomposes the predicted log-odds
of success additively,
\begin{equation}
f(\mathbf{x}_i) = \phi_0 + \sum_{j=1}^{12} \phi_{i,j},
\end{equation}
where $\phi_0$ is the baseline expected prediction. We use these attribution
vectors as behavioral signatures. Over a set of trajectories we discard any with
near-zero magnitude, for which cosine similarity is undefined,
\begin{equation}
X_{\text{filtered}} = \left\{ \boldsymbol{\phi}_i \;\middle|\; \lVert \boldsymbol{\phi}_i \rVert_2 > 10^{-9} \right\},
\end{equation}
and define the pairwise similarity between trajectories $i$ and $j$ as
\begin{equation}
S_{ij} = \frac{\boldsymbol{\phi}_i \cdot \boldsymbol{\phi}_j}{\lVert \boldsymbol{\phi}_i \rVert_2 \, \lVert \boldsymbol{\phi}_j \rVert_2}.
\end{equation}
With $k$ valid vectors, the Behavioral Consistency Metric is the mean of the
strict upper triangle of $S$,
\begin{equation}
\mathrm{BCM} = \frac{2}{k(k-1)} \sum_{i=1}^{k-1} \sum_{j=i+1}^{k} S_{ij},
\end{equation}
and is undefined for $k < 2$. The global BCM of an agent system applies this over
all of its trajectories. The within-task BCM conditions on individual task
instances: for each SWE-bench instance $t$ on which the system has at least three
trajectories, we compute $\mathrm{BCM}_t$ over that instance alone and average
across the $M$ eligible instances,
\begin{equation}
\text{within-task BCM} = \frac{1}{M} \sum_{t=1}^{M} \mathrm{BCM}_t.
\end{equation}

\subsection{Dataset Construction and Agent Cohort}
We use execution trajectories from two public Hugging Face repositories,
\texttt{SWE-bench/SWE-smith-trajectories} and
\texttt{nebius/SWE-agent-trajectories}. Both contain interaction records of
software-engineering agents on SWE-bench tasks \citep{jimenez2024swebench}, with
the open-source systems running under the SWE-agent scaffold
\citep{yang2024sweagent}; both record
command executions, file operations, test runs, intermediate outputs, and final
outcomes. The raw
corpus is roughly 24{,}000 trajectories. We remove trajectories with incomplete
logs, missing metadata, or absent termination indicators, and restrict the data
to six agent systems for direct comparison, yielding a final set of 9{,}191
trajectories. The cohort comprises 946 trajectories from
\texttt{claude-3-5-sonnet-20241022}, 2{,}938 from
\texttt{claude-3-7-sonnet-20250219}, 116 from \texttt{gpt-4o-2024-08-06}, and
1{,}191, 2{,}000, and 2{,}000 from the 405B, 70B, and 8B SWE-agent Llama
systems respectively. Each trajectory is one complete attempt to resolve one
SWE-bench issue and is the unit of analysis.

\subsection{Behavioral Feature Extraction}
Each trajectory is converted to the fixed-length representation of Equation~(1)
by parsing its ordered action sequence and aggregating action-level statistics
computable across all systems regardless of architecture. Each step is assigned to
one of four \emph{action categories} based on its command pattern---file search,
file view, file edit, or test execution---and steps matching none of these are
left uncategorized; the category-based features (including \texttt{step\_velocity},
which counts transitions between consecutive categories) are derived from this
assignment. The operational
definitions of all twelve features appear in Table~\ref{tab:features}.

\begin{table}[t]
\centering
\caption{Behavioral feature definitions.}
\label{tab:features}
\small
\begin{tabular}{ll}
\toprule
Feature & Definition \\
\midrule
total\_steps & Number of messages (steps) in the trajectory \\
mean\_action\_length & Mean character length of a message across steps \\
max\_action\_length & Maximum character length of a single message \\
file\_search\_count & Fraction of steps matching file-search command patterns \\
file\_view\_count & Fraction of steps matching file-view command patterns \\
file\_edit\_count & Fraction of steps matching file-edit command patterns \\
test\_execution\_count & Fraction of steps matching test-execution patterns \\
action\_entropy & Shannon entropy over the distribution of distinct messages \\
consecutive\_repetition\_max & Longest run of identical consecutive messages \\
unique\_action\_ratio & Ratio of distinct messages to total messages \\
error\_flag\_count & Fraction of steps matching error patterns \\
step\_velocity & Fraction of steps where the action category changes \\
\bottomrule
\end{tabular}
\end{table}

\subsection{Trajectory-Predictive Model}
\label{sec:model}
We train a supervised classifier to predict task outcome from the behavioral
features. Each trajectory carries a binary label equal to its SWE-bench
resolution status ($y=1$ for a resolved issue, $y=0$ otherwise). We use a
LightGBM classifier \citep{ke2017lightgbm} evaluated with five-fold stratified
cross-validation,
partitioning the 9{,}191 trajectories into five folds that preserve the
resolution-label distribution. Features enter the predictor at their natural
scales; gradient-boosted trees are invariant to monotone per-feature rescaling, so
no normalization is applied here (we $z$-score features only for the raw-feature
baseline of Section~5.1). For each fold $D_f$, the classifier is trained on
the remaining folds $D_{-f}$ and applied only to the held-out partition,
\begin{equation}
\hat{y}_i = M_f(\mathbf{x}_i), \qquad i \in D_f,
\end{equation}
producing a complete set of out-of-fold (OOF) predictions in which every
trajectory is scored by a model that did not train on it. The classifier attains
a mean cross-validated AUC-ROC of 0.690.

\subsection{Behavioral Attribution Vectors via SHAP}
To attribute each prediction to individual features, we compute TreeSHAP values
\citep{lundberg2017shap}
for every trajectory, yielding the attribution vector $\boldsymbol{\phi}_i$ of
Equation~(2). Attributions are produced under the same OOF scheme used for
evaluation: for each fold, SHAP values are computed for the held-out trajectories
using the model trained on the other folds. This matters for validity. Were
attributions computed from a model trained on the full dataset, the feature
contributions could partly reflect memorization and fold-specific dependencies;
restricting attribution to held-out observations ensures each trajectory is
explained by a model that did not see it. Using attribution vectors rather than
the raw features directly is deliberate: SHAP maps each feature's contribution
into a common additive log-odds space, so heterogeneous raw quantities (step
counts, character lengths, and bounded fractions) become comparable contributions
on a shared scale. This makes the downstream cosine similarity well-posed, as it
compares directions in a space where every dimension is expressed in the same
units of effect on predicted success. The resulting 12-dimensional vectors are
the basis for all consistency analysis.

\subsection{Aggregation and Uncertainty}
We compute BCM (Equation~5) under two groupings. In the global analysis,
trajectories are grouped by agent system, measuring similarity across all of a
system's trajectories. In the task-conditioned analysis, trajectories are grouped
by agent system and SWE-bench instance, BCM is computed for each instance with at
least three trajectories, and the per-system estimate averages over eligible
instances (Equation~6). Uncertainty is estimated by non-parametric bootstrap over
1{,}000 iterations: for the global BCM we resample trajectories with replacement,
and for the within-task BCM we resample the eligible task instances with
replacement, in each case recomputing the statistic and forming two-sided 95\%
confidence intervals from the 2.5th and 97.5th percentiles. A resample that leaves
fewer than two valid vectors in a group yields an undefined BCM (Equation~5) and is
discarded before computing the percentiles; this never occurs for the global
estimates and is rare for the within-task estimates, which condition on instances
with at least three trajectories. The near-zero-norm
filter of Equation~(3) is inactive on our data, removing 0 of the 9{,}191
trajectories, so per-system BCM denominators equal the full trajectory counts.

\section{Experimental Setup}
\label{sec:setup}
Our analyses compare systems along three axes: global cross-task BCM versus
success rate; a within-task control that recomputes BCM on repeated attempts at
identical instances, holding difficulty constant; and a difficulty-stratified
analysis binning tasks into easy, medium, and hard terciles by resolution rate.
The hard bin consists of tasks left unsolved under every observed execution, so we
read it as observed-unsolvable rather than an independent difficulty measure.

\section{Results}

\subsection{Consistency varies across systems and is not reducible to success}
\begin{figure}[t]
\centering
\includegraphics[width=0.72\textwidth]{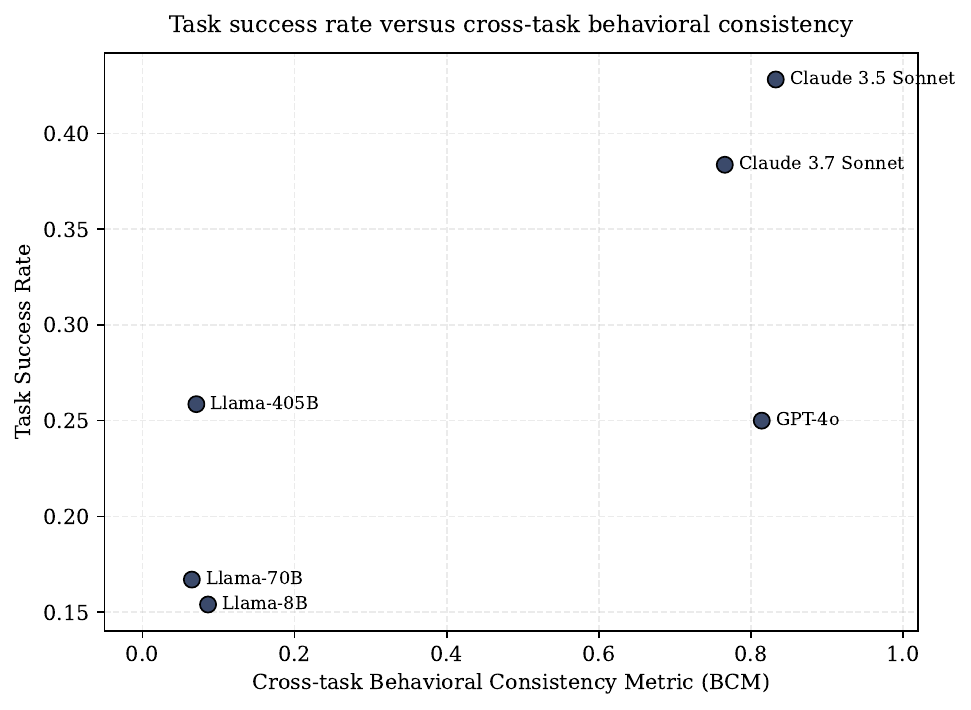}
\caption{Task success rate versus cross-task behavioral consistency (BCM) by agent system.
GPT-4o and SWE-agent Llama-405B resolve comparable fractions of tasks yet differ in
consistency by more than an order of magnitude, illustrating that consistency is not
reducible to success rate.}
\label{fig:scatter}
\end{figure}

Table~\ref{tab:global} reports global cross-task BCM for each system alongside
its success rate. Consistency varies enormously: the three proprietary API
systems occupy a high-consistency region with BCM between 0.77 and 0.83, while
the three open-source Llama SWE-agents fall between 0.065 and 0.086, more than an
order of magnitude lower. For most systems consistency and success move together,
the frontier systems being both more consistent and more successful, which is in
line with prior reports that same-task consistency tracks accuracy
\citep{mehta2026amplifies}. Consistency is nonetheless not reducible to success
rate. GPT-4o resolves 25\% of tasks with a BCM of 0.814, while the Llama-405B
SWE-agent resolves a comparable 26\% with a BCM of 0.071: comparable success,
consistency differing by more than an order of magnitude (Figure~\ref{fig:scatter}). The bootstrap intervals
for these two systems, $[0.713, 0.906]$ and $[0.056, 0.090]$, are widely
separated and do not approach overlap.

We are deliberate about how much weight this single comparison can bear. The
GPT-4o estimate rests on 116 trajectories, the smallest cohort in our study and
the source of its wide interval, and it aligns with the open-versus-closed
distinction in our sample, which, as we discuss in Section~7, is confounded with
the agent scaffold. We therefore read it not as a sweeping dissociation but as
evidence that consistency and success are separable properties, a point the
within- versus cross-task analysis below establishes on firmer ground.

To check that this separation is a property of the attribution representation
rather than of the raw features, we recompute global BCM on standardized raw
feature vectors (each feature $z$-scored) in place of SHAP vectors. The separation
collapses: standardized-raw BCM falls in a narrow $0.28$--$0.54$ band for all six
systems and shows no frontier-versus-open ordering, whereas in attribution space
the same systems span $0.07$ to $0.83$. The order-of-magnitude gap is thus visible
in attribution space but absent in raw feature space, indicating that the
success-predictive attribution, not raw feature magnitude, is what carries the
behavioral signal BCM measures.

\begin{table}[t]
\centering
\caption{Global cross-task BCM by agent system. Values are point estimates with
95\% bootstrap confidence intervals (1{,}000 resamples).}
\label{tab:global}
\begin{tabular}{lccc}
\toprule
Agent System & Global BCM & 95\% CI & Success Rate \\
\midrule
Claude 3.5 Sonnet  & 0.833 & [0.800, 0.864] & 0.43 \\
Claude 3.7 Sonnet  & 0.766 & [0.744, 0.784] & 0.38 \\
GPT-4o             & 0.814 & [0.713, 0.906] & 0.25 \\
SWE-agent Llama-405B & 0.071 & [0.056, 0.090] & 0.26 \\
SWE-agent Llama-70B  & 0.065 & [0.059, 0.073] & 0.17 \\
SWE-agent Llama-8B   & 0.086 & [0.073, 0.103] & 0.15 \\
\bottomrule
\end{tabular}
\end{table}

\subsection{The consistency gap persists under a within-task control}
A natural concern is that cross-task consistency is confounded with task
difficulty: trajectory structure varies with how hard a task is, so a system
evaluated on a more uniform slice of tasks might appear more consistent for that
reason alone. To control for this we restrict to repeated attempts on identical
task instances, where difficulty is fixed by construction, and recompute BCM
within each task before averaging. Table~\ref{tab:within} reports the result.

The frontier-versus-open-source gap survives this control. Claude 3.5 Sonnet
remains highly consistent within task (0.807, against a global 0.833), Claude 3.7
Sonnet drops but stays high (0.564), and all three Llama systems remain
substantially lower, between 0.305 and 0.348. Because difficulty is constant
within a task, the persistence of this ordering indicates that the consistency
difference reflects genuine behavioral variation rather than differing task
mixes. We are careful about scope: GPT-4o has only four task instances with three
or more attempts, far too few for a reliable within-task estimate, so this
analysis establishes the broader frontier-versus-open-source gap, not the GPT-4o
point specifically. Its within-task value is listed for completeness but should
not be interpreted. We read this frontier-versus-open gap as secondary to the
within- versus cross-task divergence and, because all open-source systems share the
SWE-agent scaffold, as confounded with the harness rather than cleanly attributable
to model openness.

\begin{table}[t]
\centering
\caption{Global versus within-task BCM. Global BCM is the point estimate;
within-task BCM averages over task instances with at least three attempts.
$^{*}$GPT-4o within-task rests on only four eligible tasks and should not be
interpreted.}
\label{tab:within}
\begin{tabular}{lcccc}
\toprule
Agent System & Global BCM & Within-Task BCM & 95\% CI & Tasks ($N \ge 3$) \\
\midrule
Claude 3.5 Sonnet  & 0.833 & 0.807 & [0.720, 0.879] & 60 \\
Claude 3.7 Sonnet  & 0.766 & 0.564 & [0.457, 0.666] & 60 \\
GPT-4o$^{*}$       & 0.814 & 0.897 & [0.819, 0.967] & 4 \\
SWE-agent Llama-405B & 0.071 & 0.348 & [0.291, 0.405] & 100 \\
SWE-agent Llama-70B  & 0.065 & 0.305 & [0.238, 0.374] & 132 \\
SWE-agent Llama-8B   & 0.086 & 0.319 & [0.272, 0.364] & 261 \\
\bottomrule
\end{tabular}
\end{table}

\subsection{Cross-task and within-task consistency are distinct axes}
\begin{figure}[t]
\centering
\includegraphics[width=0.72\textwidth]{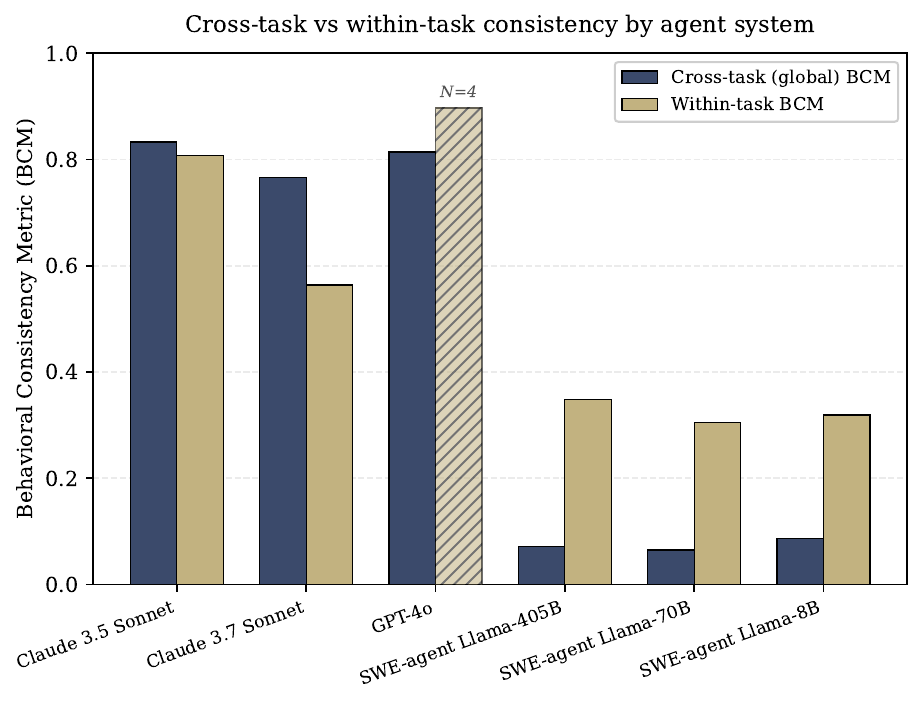}
\caption{Cross-task (global) versus within-task BCM by agent system. The
open-source Llama systems rise sharply from very low cross-task consistency to
moderate within-task consistency, while the frontier systems remain high at both
scales. GPT-4o's within-task bar (hatched) rests on only four eligible tasks and
should not be interpreted.}
\label{fig:globalwithin}
\end{figure}

The within-task numbers reveal our central result, and unlike the single-system
comparison above it rests on all three open-source systems with large samples,
between 100 and 261 task instances each. For the open-source systems, within-task
consistency is markedly higher than cross-task consistency (Figure~\ref{fig:globalwithin}): Llama-405B rises from
a global 0.071 to 0.348 within task, and the other two Llama systems show the same
pattern. When these agents attempt the same task more than once they behave
fairly similarly, but their behavior fragments when they move across different
tasks. The frontier systems do not show this gap to nearly the same degree;
Claude 3.5 Sonnet is about as consistent within task as across tasks. The
bootstrap intervals in Table~\ref{tab:within} confirm the ordering is not an
artifact of task sampling: Claude 3.7 Sonnet's within-task interval $[0.457,
0.666]$ lies entirely above every open-source interval (all upper bounds
$\le 0.405$).

Cross-task and within-task consistency are therefore not the same property at
different resolutions. They can move in opposite directions: an agent can be
locally reproducible yet globally unstable, repeating its approach on a fixed task
while carrying no consistent strategy from one task to the next. This is the
distinction that separates our measure from the same-task reproducibility metrics
of Section~2, which by construction cannot detect global fragmentation in a
locally consistent agent. We regard it as the most informative of our findings,
because it identifies a specific profile, reproducible locally and incoherent
globally, that both outcome metrics and same-task consistency metrics miss.

\subsection{Behavioral archetypes}
\begin{figure}[t]
\centering
\includegraphics[width=0.72\textwidth]{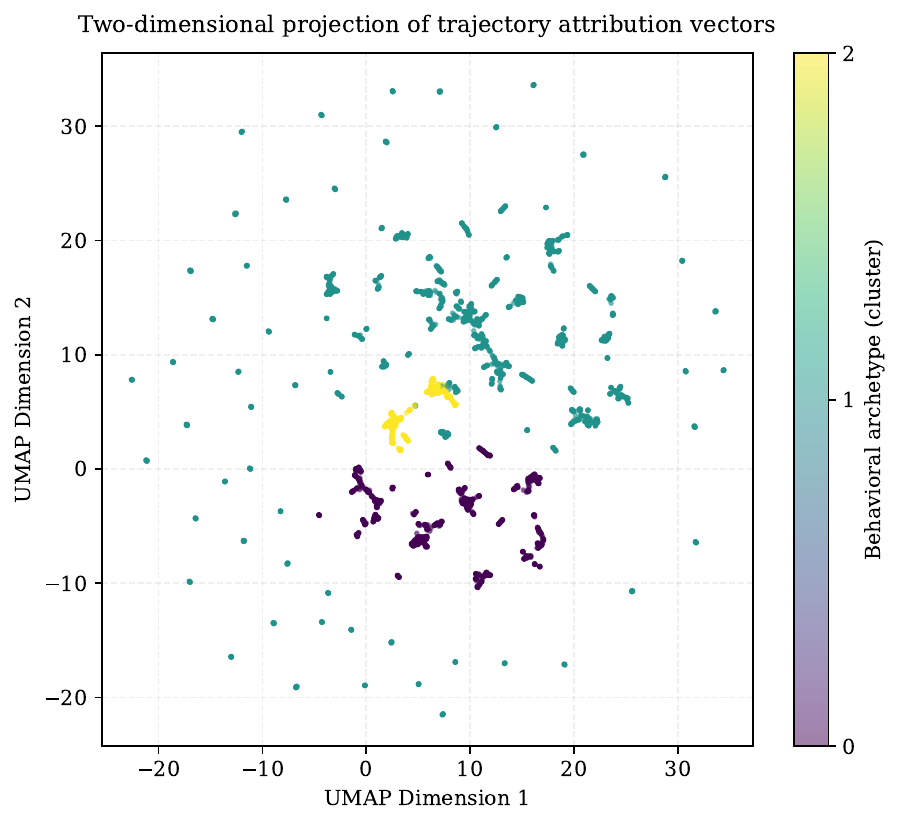}
\caption{Two-dimensional UMAP projection of trajectory attribution vectors, colored by
behavioral archetype ($k=3$). Frontier systems concentrate in the \emph{decisive executor}
cluster (top); open-source Llama systems distribute across all three, with the dominant
archetype varying by task instance.}
\label{fig:umap}
\end{figure}

To characterize these behavioral profiles, we cluster the attribution vectors
with $k$-means and project them to two dimensions using UMAP
(Figure~\ref{fig:umap}). We treat the clustering as descriptive: a silhouette analysis over
the attribution vectors favors $k=2$ (0.58) with $k=3$ close behind (0.49), and we
use $k=3$ to surface a third behavioral mode rather than to claim a uniquely
correct partition. Silhouette values in this range indicate soft, overlapping
cluster boundaries rather than well-separated groups, so the archetypes below
should be read as exploratory rather than as a definitive partition of the
trajectory space. The trajectories separate into three descriptive archetypes.
The first, \emph{decisive executors}, consists of short, low-error trajectories
that resolve tasks efficiently (mean length 21 steps, success rate 0.35). The
second, \emph{drifting explorers}, comprises long, high-error trajectories that
take many steps without resolving (mean length 104 steps, success rate 0.07). The
third, \emph{verbose minimalists}, consists of short trajectories built from very
long, verbose individual actions (success rate 0.18). The link to BCM is one of
concentration versus scatter: the high-consistency frontier systems concentrate in
a single archetype (decisive executors, 61\% frontier), whereas the
low-consistency open-source systems distribute across all three, with the
dominant archetype shifting from one task instance to the next. A system
concentrated in one archetype produces similar signatures and scores high on
cross-task consistency; a system scattered across archetypes produces dissimilar
signatures and scores low.

\subsection{Consistency across difficulty}
\begin{figure}[t]
\centering
\includegraphics[width=0.72\textwidth]{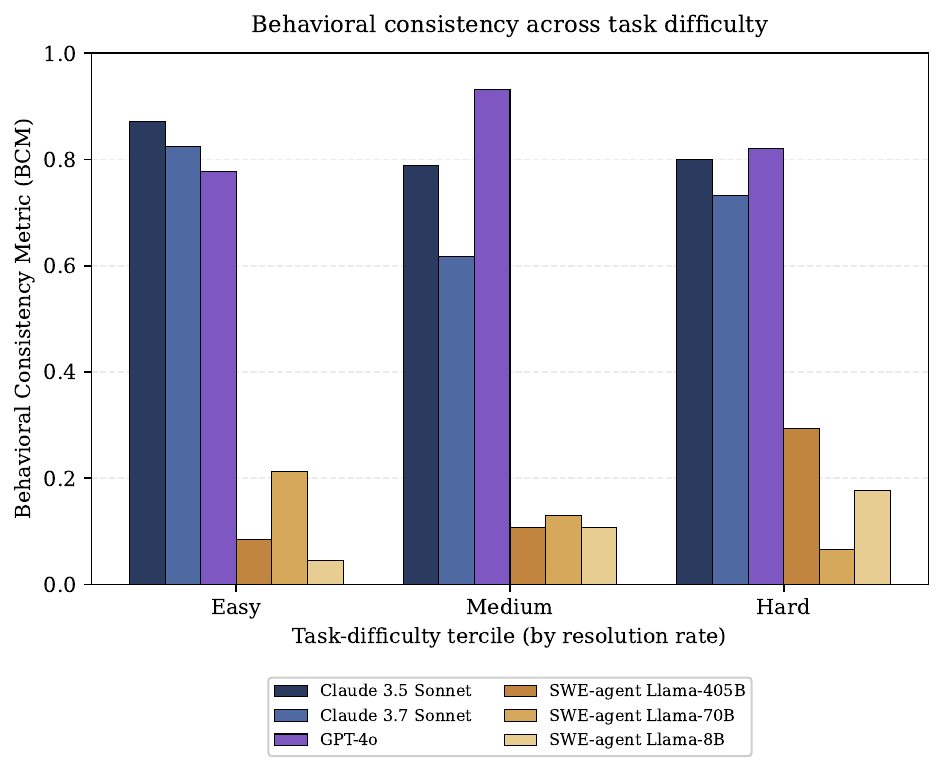}
\caption{Behavioral consistency (BCM) by task-difficulty tercile. Frontier systems remain
roughly flat across all three bins. Open-source systems vary without a consistent direction.
The hard bin consists entirely of observed-unsolvable tasks and should be read descriptively.}
\label{fig:difficulty}
\end{figure}

Finally, we examine how consistency varies with difficulty, binning tasks into
terciles by overall resolution rate (Table~\ref{tab:difficulty}, Figure~\ref{fig:difficulty}). The frontier
systems are roughly stable across bins, maintaining high consistency on easy,
medium, and hard tasks alike. The open-source systems are more variable and do
not move in a single direction: Llama-405B becomes more consistent on harder
tasks while Llama-70B becomes less so. We read these results cautiously. As noted
in Section~\ref{sec:setup}, the hard bin consists of tasks unsolved under every observed
execution, so it reflects observed unsolvability rather than an independent
difficulty measure. We therefore report the difficulty-stratified pattern as
descriptive and draw no predictive or early-warning claim from it.

\begin{table}[t]
\centering
\caption{Behavioral consistency (BCM) by task-difficulty tercile.}
\label{tab:difficulty}
\begin{tabular}{lccc}
\toprule
Agent System & Easy & Medium & Hard \\
\midrule
Claude 3.5 Sonnet  & 0.871 & 0.789 & 0.800 \\
Claude 3.7 Sonnet  & 0.824 & 0.618 & 0.733 \\
GPT-4o             & 0.778 & 0.931 & 0.821 \\
SWE-agent Llama-405B & 0.085 & 0.107 & 0.293 \\
SWE-agent Llama-70B  & 0.213 & 0.130 & 0.066 \\
SWE-agent Llama-8B   & 0.045 & 0.108 & 0.177 \\
\bottomrule
\end{tabular}
\end{table}

\section{Discussion}

Our results suggest that behavioral consistency is a property worth measuring in
its own right, separate from how often an agent succeeds. We read BCM not as a
competitor to success rate but as a complement: success rate answers whether an
agent tends to reach the goal, and BCM answers whether it gets there the same way
each time. The value of the measure is concentrated in the cases where success
rate is held roughly fixed and consistency varies, since there the two metrics
genuinely come apart and the consistency signal adds information the outcome
cannot. A high score is not on its own a mark of quality, though: an agent that
fails the same way on every task is perfectly consistent and entirely unreliable,
so BCM is only meaningful read alongside competence.

The separation between cross-task and within-task consistency is our central
result. An agent stable when it retries the same task but scattered across
different tasks has reproducible local behavior but no transferable global
strategy, exactly the distinction that same-task reproducibility cannot capture.
It also opens an interpretive question we do not resolve, namely what a stable
cross-task strategy actually consists of beyond the aggregate signature we
measure.

Finally, the measure as presented is diagnostic rather than predictive: it
characterizes an agent after its runs are complete. A natural next step is to ask
whether the same behavioral signals observed partway through a run can anticipate
its outcome early enough to act on, turning a descriptive measure into a usable
reliability monitor. We frame this as a hypothesis rather than a result, since
establishing it would require showing that early behavioral signals beat the
obvious length-based and progress-based baselines.

\section{Limitations}
\label{sec:limitations}
BCM should be read as a measure of the consistency of \emph{success-predictive
structural trajectory signatures}, which we treat as a measurable proxy for
behavioral consistency rather than a direct measure of semantic strategy: the
twelve features are structural (step counts, edits, errors, repetition) and do not
capture whether an agent formed the right hypothesis or made relevant edits, so our
use of the word ``strategy'' is informal. It inherits its representation from a
single success predictor (cross-validated AUC 0.690) and its TreeSHAP attributions;
we did not test sensitivity to the predictor family, the attribution method, the
feature set, or the similarity measure, and establishing robustness to these
choices is the most important validation we leave open. The \emph{ordering} of
systems is more likely to be robust than the \emph{magnitude} of the separation:
our standardized-raw-feature baseline (Section~5.1) already shows the
order-of-magnitude gap is a property of the attribution representation rather than
the raw features, so the absolute BCM values should not be over-interpreted.
Because the predictor is trained across systems, part of the between-system
separation could also reflect system-correlated feature use; isolating
success-relevant signal from predictor-encoded system identity would require a
label-permutation or per-system control we do not run. We ran no temperature ablation, so low consistency could
partly reflect base-model sampling variation. The within-task estimates condition
on instances with at least three attempts, which may over-represent repeatedly
attempted tasks. The open-versus-closed comparison is confounded by the harness,
since all open-source agents share the SWE-agent infrastructure. Validation covers
one task domain, six configurations, and one benchmark family, so generalization
is unestablished, and the GPT-4o single-system dissociation rests on only 116
trajectories; the within- versus cross-task result, resting on all three
open-source systems with large samples, does not depend on it.

\section{Conclusion}
We introduced the Behavioral Consistency Metric to make cross-task behavioral
consistency measurable from execution traces. Across six systems on software
engineering tasks, cross-task and within-task consistency emerge as distinct axes
that can diverge, the frontier-versus-open-source gap survives a within-task
control, and consistency is separable from success rate, a reliability dimension
that outcome metrics alone leave invisible.

\section{Acknowledgements}
The authors would like to thank Professor Manolis Kellis for his guidance, feedback, and mentorship throughout this project. The authors also thank the Stanford AI Measurement Science (AIMS) Laboratory for their valuable feedback and suggestions for the Conference on Language Modeling's (COLM) AIMS workshop submission.

\section*{Data and Code Availability}
The trajectory data, feature-extraction and attribution pipeline, and all code
required to reproduce the metrics, tables, and figures in this paper are available
at \url{https://github.com/pranilraichura/colm-agent-eval}. The underlying
execution traces are drawn from the public \texttt{SWE-bench/SWE-smith-trajectories}
and \texttt{nebius/SWE-agent-trajectories} datasets on Hugging Face.

\bibliographystyle{abbrvnat}
\bibliography{references}

\end{document}